\documentclass[11pt]{article}
\usepackage{naaclhlt2015}
\usepackage{times}
\usepackage{latexsym}

\usepackage{examples}
\usepackage{balance}
\usepackage{cgloss4e}
\usepackage{verbatim}
\usepackage{cprotect}

\usepackage{listings}
\usepackage{url}
\usepackage{amsmath}
\usepackage{stackrel}
\usepackage[pdftex]{graphicx}
\usepackage{dblfloatfix}
\usepackage{graphics}
\usepackage{array}
\newcolumntype{x}[1]{>{\centering\arraybackslash\hspace{0pt}}p{#1}}
\newcolumntype{y}[1]{<{\hspace{-.15cm}}p{#1}}

\usepackage[usenames,dvipsnames,svgnames,table]{xcolor}

\usepackage{amssymb,amsmath,epsfig}
\usepackage{mathpartir}
\usepackage{proof}
\usepackage{amsthm}
\usepackage{xspace}
\usepackage{algorithm}
\usepackage[noend]{algpseudocode}
\algrenewcommand{\algorithmicindent}{1em}

\usepackage{enumitem}

\newcommand{\notes}[1]{}

 \theoremstyle{definition}
 
\theoremstyle{plain}

\newcommand{\ith}[1]{\ensuremath{i^{{th}}}}

\newcount\permx
\newcount\permy
\def\permdot#1#2{
\permx=#1 \advance\permx by-1
\permy=#2 \advance\permy by-1
\psframe[fillcolor=black, fillstyle=solid]
(\permx,\permy)(#1, #2)
}

\newcommand{\boxnum}[1]{{\setlength{\fboxsep}{1pt}\raisebox{1pt}{\hspace{1pt}\fbox{\tiny #1}\hspace{1pt}}}}
\newcommand{\ind}[1]{\ensuremath{_{\kern-0.5pt\boxnum{#1}}}}

\def\namecite{\newcite}

\newcommand{\smallnt}[1]{\ensuremath{_{\mbox{\tiny PP}}}\xspace}

\newcommand{\pseudocode}{Algorithm}
\floatname{algorithm}{\pseudocode}

\iffalse

\else

\fi

\newcommand{\ter}{{\sc Ter}\xspace}
\newcommand{\bleu}{{\sc Bleu}\xspace}
\newcommand{\tb}{{\sc T-b}\xspace}
\newcommand{\bp}{{\sc Bp}\xspace}

\usepackage{multirow}
\usepackage{tikz}
\usetikzlibrary{positioning}
\usepackage{tikz-qtree}
\usetikzlibrary{arrows}

\title{Coverage Embedding Models for Neural Machine Translation}

\iftrue
\author{
Haitao Mi \;\;  Baskaran Sankaran \;\; Zhiguo Wang  \;\; Abe Ittycheriah\thanks{\;\;Work done while at IBM.  To contact Abe, aittycheriah@google.com.}\\
T.J.~Watson Research Center \\
IBM \\
1101 Kitchawan Rd, Yorktown Heights, NY 10598 \\
{\tt \{hmi, bsankara, zhigwang, abei\}@us.ibm.com}
}
\fi
\date{}

\begin{document}
\maketitle

\begin{abstract}
In this paper, we enhance the attention-based neural machine translation (NMT)
by adding explicit coverage embedding models to alleviate 
issues of repeating and dropping translations in NMT.
For each source word, our model starts with a {\em full} coverage embedding vector
to track the {\em coverage} status,
and then keeps updating it with neural networks as the translation goes.
Experiments on the large-scale Chinese-to-English task
show that our enhanced model improves the translation quality 
significantly on various test sets over the strong large vocabulary NMT system.

\end{abstract}

\setlength{\belowdisplayskip}{3pt} \setlength{\belowdisplayshortskip}{3pt}
\setlength{\abovedisplayskip}{3pt} \setlength{\abovedisplayshortskip}{3pt}
\section{Introduction}
\label{sec:intro}
Neural machine translation (NMT) has gained popularity in recent years 
(e.g. \cite{bahdanau+:2014,jean+:2015,luong+:2015,mi+:2016,li+:2016}),
especially for the attention-based models of \namecite{bahdanau+:2014}.
The attention at each time step shows which source word the 
model should focus on to predict the next target word.
However, the attention in each step only looks at the previous hidden state
and the previous target word, 
there is no history or coverage information typically for each source word. 
As a result, this kind of model suffers from issues of repeating or dropping translations.

The traditional statistical machine translation (SMT) 
systems (e.g.~\cite{koehn:2004})
address the above issues by employing a source side ``coverage vector'' for each sentence 
to indicate explicitly which words have been translated, which parts have not yet.
A coverage vector starts with all zeros, meaning no word has been translated.
If a source word at position $j$ got translated, the coverage vector sets 
position $j$ as 1, and they won't use this source word in future translation.
This mechanism avoids the repeating or dropping translation problems.

However, it is not easy to adapt the ``coverage vector'' to NMT directly,
as attentions are soft probabilities, not 0 or 1. 
And SMT approaches handle one-to-many fertilities by using 
phrases or hiero rules (predict several words in one step),
while NMT systems only predict one word at each step.

In order to alleviate all those issues,
we borrow the basic idea of ``coverage vector'', 
and introduce a coverage embedding vector for each source word. 
We keep updating those embedding vectors at each translation step,
and use those vectors to track the {\em coverage} information.

Here is a brief description of our approach.
At the beginning of translation, 
we start from a {\em full} coverage embedding vector for each source word.
This is different from the ``coverage vector'' in SMT in following two aspects: 
\begin{enumerate}[topsep=0pt,itemsep=-1ex,partopsep=1ex,parsep=1ex]
\item[$\bullet$] each source word has its own coverage embedding vector, instead of 0 or 1, a scalar, in SMT,
\item[$\bullet$] we start with a {\em full} embedding vector for each word, instead of 0 in SMT.
\end{enumerate}
After we predict a translation word $y_t$ at time step $t$, 
we need to update each coverage embedding vector accordingly 
based on the attentions 
in the current step.
Our motivation 
is that if we observe a very high attention over $x_i$ in this step,
there is a high chance that $x_i$ and $y_t$ are translation equivalent.
So the embedding vector of $x_i$ should come to {\em empty} (a zero vector) in a one-to-one translation case,
or subtract the embedding of $y_t$ for the one-to-many translation case. 
An {\em empty} coverage embedding of a word $x_i$ indicates this word is translated, and we can not translate $x_i$ again in future.
Empirically, we model this procedure by using neural networks
(gated recurrent unit (GRU)~\cite{gru} or direct subtraction). 

Large-scale experiments over Chinese-to-English on 
various test sets show that
our method improves the translation quality significantly 
over the large vocabulary NMT system (Section~\ref{sec:exps}).

\begin{figure*}[!t]
\centering
\includegraphics[width=0.9\textwidth]{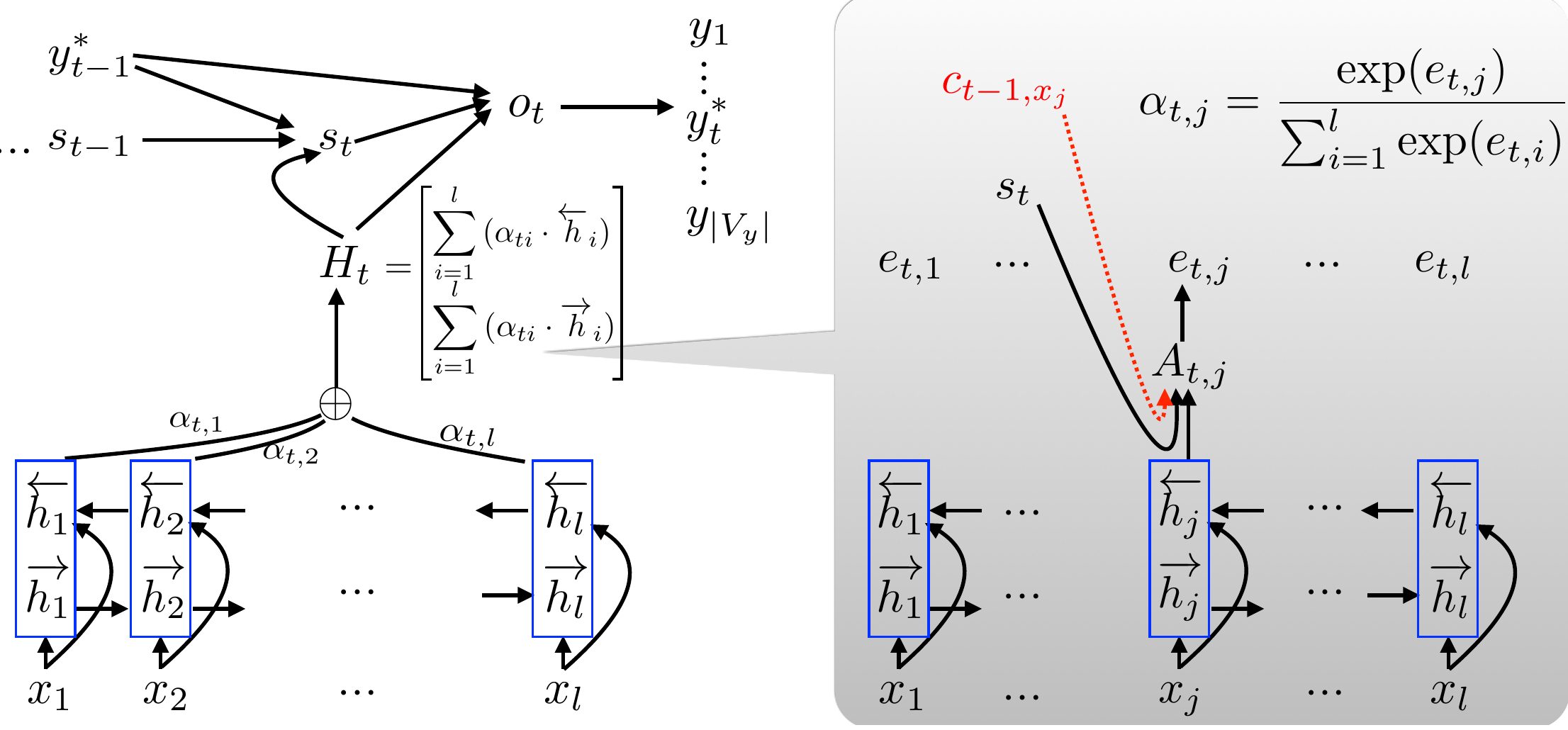}
\vspace{-0.3cm}
\caption{The architecture of attention-based NMT.
The source sentence is ${\bf{x}}=({x_1, ... , x_l})$ with length $l$, 
the translation is ${\bf{y^*}}=({y^*_1, ... , y^*_m})$ with length $m$.
$\overleftarrow{h_i}$ and $\overrightarrow{h_i}$ are bi-directional encoder states.
$\alpha_{t,j}$ is the attention probability at time $t$, position $j$.
$H_{t}$ is the weighted sum of encoding states.
$s_t$ is a hidden state. $o_t$ is an output state. 
Another one layer neural network projects $o_t$ to the target output vocabulary,
and conducts softmax to predict the probability distribution over the output vocabulary.
The attention model (in right gray box) is a two layer feedforward neural network, 
$A_{t,j}$ is an intermediate state, then another layer converts 
it into a real number $e_{t,j}$, the final attention probability
at position $j$ is $\alpha_{t,j}$.
We plug coverage embedding models into NMT model by adding an input $c_{t-1,x_j}$ to $A_{t,j}$ 
(the red dotted line).
}
\label{fig:att}
\vspace{-0.5cm}
\end{figure*}

\section{Neural Machine Translation}
\label{sec:nmt}
As shown in Figure~\ref{fig:att}, 
attention-based neural machine translation \cite{bahdanau+:2014} 
is an encoder-decoder network.  
the encoder employs a bi-directional recurrent neural network to 
encode the source sentence ${\bf{x}}=({x_1, ... , x_l})$, 
where $l$ is the sentence length, into
a sequence of hidden states ${\bf{h}}=({h_1, ..., h_l})$,
each $h_i$ is a concatenation of a left-to-right $\overrightarrow{h_i}$
and a right-to-left $\overleftarrow{h_i}$,
\[
h_{i} = 
\begin{bmatrix}
\overleftarrow{h}_i \\ 
\overrightarrow{h}_i \\
\end{bmatrix}
=
\begin{bmatrix}
\overleftarrow{f}(x_i, \overleftarrow{h}_{i+1}) \\
\overrightarrow{f}(x_i, \overrightarrow{h}_{i-1}) \\
\end{bmatrix},
\]
where $\overleftarrow{f}$ and $\overrightarrow{f}$ 
are two GRUs.

Given the encoded ${\bf h}$, the decoder predicts the target translation
by maximizing the conditional log-probability of the 
correct translation ${\bf y^*} = (y^*_1, ... y^*_m)$, where 
$m$ is the sentence length. At each time $t$, 
the probability of each word $y_t$ from a target vocabulary $V_y$ is:
\begin{equation}
\label{eq:py}
p(y_t|{\bf h}, y^*_{t-1}..y^*_1) = g(s_t, y^*_{t-1}),
\end{equation}
where $g$ is 
a two layer feed-forward neural network ($o_t$ is a intermediate state) over
the embedding of the previous word $y^*_{t-1}$, and  
the hidden state $s_t$. 
The $s_t$ is computed as:
\begin{equation}
s_t = q(s_{t-1}, y^*_{t-1}, H_{t})
\end{equation}
\begin{equation}
H_t = 
\begin{bmatrix}
\sum_{i=1}^{l}{(\alpha_{t,i} \cdot \overleftarrow{h}_i)} \\
\sum_{i=1}^{l}{(\alpha_{t,i} \cdot \overrightarrow{h}_i)} \\
\end{bmatrix},
\end{equation}
where $q$ is a GRU, $H_t$ is a weighted sum of ${\bf h}$,
the weights, $\alpha$, are computed with a two layer feed-forward neural network $r$:
\begin{equation}
\alpha_{t,i} = \frac{\exp\{r(s_{t-1}, h_{i}, y^*_{t-1})\}}{\sum_{k=1}^{l}{\exp\{r(s_{t-1}, h_{k}, y^*_{t-1})\}}}
\end{equation}

\section{Coverage Embedding Models}
\label{sec:cov}
Our basic idea is to introduce a coverage embedding for each source word, 
and keep updating this embedding at each time step.
Thus, the coverage embedding for a sentence is a matrix, instead of a vector in SMT.
As different words have different fertilities (one-to-one, one-to-many, or one-to-zero),
similar to word embeddings, each source word has its own coverage embedding vector. 
For simplicity, the number of coverage embedding vectors is the same as 
the source word vocabulary size. 

At the beginning of our translation, 
our coverage embedding matrix ($c_{0,x_1}, c_{0,x_2}, ... c_{0,x_l}$) 
is initialized with the coverage embedding vectors of all the source words.

Then we update them with neural networks (a GRU (Section~\ref{sec:gru}) or a subtraction (Section~\ref{sec:sub})) 
until we translation all the source words.

In the middle of translation, 
some coverage embeddings should be close to {\em zero}, which indicate those words are covered or translated,
and can not be translated in future steps.
Thus, in the end of translation, the embedding matrix should be close to zero,
which means all the words are covered.

In the following part, we first show two updating methods, 
then we list the NMT objective that takes into account the embedding models.

\subsection{Updating Methods}
\subsubsection{Updating with a GRU}
\label{sec:gru}
\begin{figure}[!t]
\centering
\includegraphics[width=0.4\textwidth]{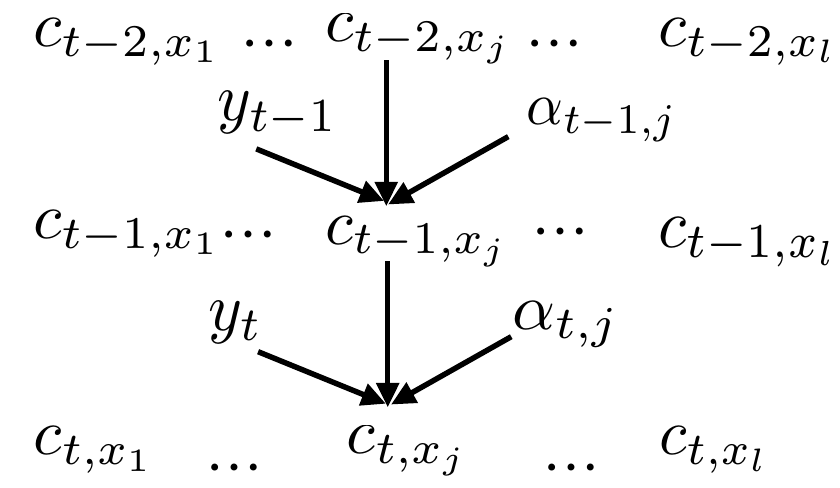}
\vspace{-0.5cm}
\caption{The coverage embedding model with a GRU at time step $t-1$ and $t$. 
$c_{0,1}$ to $c_{0,l}$ are initialized with the word coverage embedding matrix}
\label{fig:cov}
\end{figure}

Figure~\ref{fig:cov} shows the updating method with a GRU.
Then, at time step $t$, 
we feed $y_{t}$ and $\alpha_{t,j}$ to the coverage model (shown in Figure~\ref{fig:cov}), 
\begin{equation*}
\begin{split}
& z_{t,j} = \sigma(W^{zy}y_{t} + W^{z\alpha}\alpha_{t,j} + U^{z}c_{t-1,x_j}) \\
& r_{t,j} = \sigma(W^{ry}y_{t} + W^{r\alpha}\alpha_{t,j} + U^{r}c_{t-1,x_j}) \\
& \tilde{c}_{t,x_j} = \tanh(W y_{t} + W^{\alpha} \alpha_{t,j} + r_{t,j} \circ U c_{t-1,x_j})\\
& c_{t,x_j} = z_{t,j} \circ c_{t-1, x_j} + (1-z_{t,j}) \circ \tilde{c}_{t,x_j}, \\
\end{split}
\end{equation*}
where, $z_t$ is the update gate, $r_t$ is the reset gate, 
$\tilde{c}_t$ is the new memory content, and $c_t$ is the final memory.
The matrix $W^{zy}$, $W^{z\alpha}$, $U^{z}$, $W^{ry}$, $W^{r\alpha}$, $U^{r}$,
$W^{y}$, $W^\alpha$ and $U$ are shared across different position $j$.
$\circ$ is a pointwise operation. 

\subsubsection{Updating as Subtraction}
\label{sec:sub}
Another updating method is to subtract the embedding of $y_t$ directly from the coverage embedding $c_{t,x_j}$
with a weight $\alpha_{t, j}$ as
\begin{equation}
c_{t, x_j} = c_{t-1, x_j} - \alpha_{t, j} \circ (W^{y \rightarrow c}y_t),
\end{equation}
where $W^{y \rightarrow c}$ is a matrix that coverts word embedding of $y_t$ to 
the same size of our coverage embedding vector $c$.

\subsection{Objectives}
We integrate our coverage embedding models into the attention NMT~\cite{bahdanau+:2014} 
by adding $c_{t-1, x_j}$ to the first layer of the attention model (shown in the red dotted line in Figure~\ref{fig:att}).

Hopefully, if $y_t$ is partial translation of $x_j$ with a probability $\alpha_{t,j}$, 
we only remove partial information of $c_{t-1,x_j}$. 
In this way, we enable coverage embedding $c_{0,x_j}$ to encode fertility information
of $x_j$.

As we have mentioned, in the end of translation, 
we want all the coverage embedding vectors to be close to {\em zero}.
So we also minimize the absolute values of embedding matrixes as
\begin{equation}
\begin{split}
\theta^* = \arg\max_{\theta}{\sum_{n=1}^{N}} & \Bigg\{{\sum_{t=1}^{m}{\log p(y^{*n}_t|{\bf x}^n, y^{*n}_{t-1}..y^{*n}_1})} \\
      & - \lambda {\sum_{i=1}^{l}||c_{m,x_i}||} \Bigg\},
\end{split}
\label{eq:mix}
\end{equation}
where $\lambda$ is the coefficient of our coverage model.

As suggested by \namecite{mi+alignment:2016}, we can also use some supervised alignments in our training.
Then, we know exactly when each $c_{t, x_j}$ should become close to {\em zero} after step $t$. 
Thus, we redefine Equation~\ref{eq:mix} as:
\begin{equation}
\begin{split}
\theta^* = \arg\max_{\theta}{\sum_{n=1}^{N}} & \Bigg\{{\sum_{t=1}^{m}{\log p(y^{*n}_t|{\bf x}^n, y^{*n}_{t-1}..y^{*n}_1})} \\
      & - \lambda {\sum_{i=1}^{l}(\sum_{j=a_{x_i}}^m||c_{j,x_i}||)} \Bigg\},
\end{split}
\label{eq:align_obj}
\end{equation}
where $a_{x_i}$ is the maximum index on the target sentence $x_i$ can be aligned to.

\section{Related Work}
\label{sec:related}
There are several parallel and independent related work~\cite{tu+:2016,feng+:2016,cohn+:2016}. 
\namecite{tu+:2016} is the most relevant one.
In their paper, they also employ a GRU to model the coverage vector. 
One main difference is that 
our model introduces a specific coverage embedding vector for each source word,
in contrast, their work initializes the word coverage vector with a scalar with a uniform distribution.
Another difference lays in the fertility part, 
\namecite{tu+:2016} add an accumulate operation and a fertility function to simulate 
the process of one-to-many alignments. In our approach, we add fertility information 
directly to coverage embeddings, as each source word has its own embedding.
The last difference is that 
our baseline system~\cite{mi+:2016} is an extension of the large vocabulary NMT of \namecite{jean+:2015} 
with candidate list decoding
and UNK replacement, a much stronger baseline system. 

\namecite{cohn+:2016} augment the attention model with well-known features in 
traditional SMT, including positional bias, Markov conditioning, 
fertility and agreement over translation directions. This work is orthogonal to our work.

\section{Experiments}
\label{sec:exps}

\subsection{Data Preparation}
We run our experiments on Chinese to English task.
We train our machine translation systems on two training sets.
The first training corpus consists of approximately 5 million 
sentences available within the DARPA BOLT Chinese-English task. 
The second training corpus adds HK Law, HK Hansard and UN data, 
the total number of training sentence pairs is 11 million.
The Chinese text is segmented with a segmenter trained on CTB data 
using conditional random fields (CRF). 

Our development set is the concatenation of several tuning sets 
(GALE Dev, P1R6 Dev, and Dev 12)
released under the DARPA GALE program.
The development set is 4491 sentences in total.
Our test sets are NIST MT06, 
MT08 news, 
and MT08 web.

\begin{table*}[t]
\centering
\tabcolsep=0.15cm
\begin{tabular}{cc||ccc|ccc|ccc||c}
\multicolumn{2}{c||}{\multirow{3}{*}{single system}} & \multicolumn{3}{c|}{\multirow{2}{*}{MT06}}  & \multicolumn{6}{c||}{MT08} & \multirow{2}{*}{avg.}\\
                                      &              &  & & & \multicolumn{3}{c|}{News} & \multicolumn{3}{c||}{Web} &  \\\cline{3-12}
                                      &              & \bp & \bleu & \tb & \bp & \bleu & \tb & \bp & \bleu & \tb & \tb \\ 
\hline
\multicolumn{2}{c||}{Tree-to-string} & 0.95 & 34.93 & 9.45 & 0.94 & 31.12 & 12.90 & 0.90 & 23.45 & 17.72 & 13.36 \\
\hline \hline
\multicolumn{2}{c||}{LVNMT}          & 0.96 & 34.53 &12.25 & 0.93 & 28.86 & 17.40 & 0.97 & 26.78 & 17.57 & 15.74 \\
\hline
\parbox[t]{2mm}{\multirow{4}{*}{\rotatebox[origin=c]{90}{Ours}}}  & {\bf U$_{GRU}$} & 0.92 & 35.59 & 10.71 & 0.89 & 30.18 & 15.33 & 0.97 & 27.48 & 16.67 & 14.24 \\
& {\bf U$_{Sub}$}                 & 0.91 & 35.90 & 10.29 & 0.88 & 30.49 & 15.23 & 0.96 & 27.63 & 16.12 & 13.88 \\
& {\bf U$_{GRU}$}+{\bf U$_{Sub}$} & 0.92 & 36.60 &  9.36 & 0.89 & 31.86 & 13.69 & 0.95 & 27.12 & 16.37 & 13.14 \\
& +{\bf Obj.}                     & 0.93 & 36.80 &  9.78 & 0.90 & 31.83 & 14.20 & 0.95 & 28.28 & 15.73 & 13.24 \\
\hline
\end{tabular}
\caption{Single system results in terms of (\ter-\bleu)/2 (the lower the better) 
on 5 million Chinese to English training set. 
NMT results are on a large vocabulary ($300k$) and with UNK replaced. 
{\bf U$_{GRU}$}: updating with a GRU;
{\bf U$_{Sub}$}: updating as a subtraction;
{\bf U$_{GRU}$} + {\bf U$_{Sub}$}: combination of two methods (do not share coverage embedding vectors);
{\bf +Obj.}: {\bf U$_{GRU}$} + {\bf U$_{Sub}$} with an additional objective in Equation~\ref{eq:mix}, 
we have two $\lambda$s for {\bf U$_{GRU}$} and {\bf U$_{Sub}$} separately, 
and we test $\lambda_{GRU}$ = $1 \times 10^{-4}$ and $\lambda_{Sub}$ = $1 \times 10^{-2}$.
\label{tab:zhen}}
\vspace{-0.3cm}
\end{table*}

For all NMT systems, the full vocabulary sizes for thr two training sets are $300k$
and $500k$ respectively.
The coverage embedding vector size is 100.
In the training procedure, we use AdaDelta~\cite{adadelta} to 
update model parameters with a mini-batch size 80.
Following \namecite{mi+:2016},
the output vocabulary for each mini-batch or sentence is a sub-set of the full vocabulary.
For each source sentence, the sentence-level target vocabularies are
union of top $2k$ most frequent target words and the top 10 candidates of the word-to-word/phrase translation tables
learned from `fast\_align'~\cite{dyer+:2013}. 
The maximum length of a source phrase is 4.
In the training time, we add the reference in order to make the translation reachable.

Following \namecite{jean+:2015}, 
We dump the alignments, attentions, for each sentence, and 
replace UNKs with the word-to-word translation model or the aligned source word.

Our traditional SMT system is a hybrid syntax-based tree-to-string model~\cite{zhao+yaser:2008}, 
a simplified version of \namecite{liu+:2009} and \namecite{cmejrek+:2013}.
We parse the Chinese side with Berkeley parser, and align the bilingual sentences with GIZA++.
Then we extract Hiero and tree-to-string rules on the training set.
Our two 5-gram language models are trained on the English side of the parallel corpus, 
and on monolingual corpora (around 10 billion words from Gigaword (LDC2011T07)), respectively.
As suggestion by \namecite{zhang:2016},
NMT systems can achieve better results with the help of those monolingual corpora.
We tune our system with PRO~\cite{hopkins+may:2011}
to minimize (\ter - \bleu)/2 on the development set.

\subsection{Translation Results}

Table~\ref{tab:zhen} shows the results of all systems on 5 million training set. 
The traditional syntax-based system achieves 9.45, 12.90, and 17.72
on MT06, MT08 News, and MT08 Web sets respectively, and 13.36 on average in terms of (\ter - \bleu)/2.
The large-vocabulary NMT (LVNMT), our baseline, achieves an average (\ter - \bleu)/2 score of 15.74, 
which is about 2 points worse than the hybrid system.

We test four different settings for our coverage embedding models:
\begin{enumerate}[topsep=0pt,itemsep=-1ex,partopsep=1ex,parsep=1ex]
\item[$\bullet$] {\bf U$_{GRU}$}: updating with a GRU;
\item[$\bullet$] {\bf U$_{Sub}$}: updating as a subtraction;
\item[$\bullet$] {\bf U$_{GRU}$} + {\bf U$_{Sub}$}: combination of two methods (do not share coverage embedding vectors);
\item[$\bullet$] {\bf +Obj.}: {\bf U$_{GRU}$} + {\bf U$_{Sub}$} plus an additional objective in Equation~\ref{eq:mix}\footnote{We use two $\lambda$s for {\bf U$_{GRU}$} and {\bf U$_{Sub}$} separately, 
and we test $\lambda_{GRU}$ = $1\times 10^{-4}$ and $\lambda_{Sub}$ = $1 \times 10^{-2}$ in our experiments.}.
\end{enumerate}

{\bf U$_{GRU}$} improves the translation quality by 1.3 points on average over LVNMT.
And {\bf U$_{GRU}$} + {\bf U$_{Sub}$} achieves the best average score of 13.14,
which is about 2.6 points better than LVNMT. 
All the improvements of our coverage embedding models over LVNMT
are statistically significant 
with the sign-test of~\namecite{collins+:2005}.
We believe that we need to explore more hyper-parameters of {\bf +Obj.} in order to 
get even better results over {\bf U$_{GRU}$} + {\bf U$_{Sub}$}.

\begin{table}[t]
\centering
\tabcolsep=0.05cm
\small
\begin{tabular}{cc||cc|cc|cc||c}
\multicolumn{2}{c||}{\multirow{3}{*}{single system}} & \multicolumn{2}{c|}{\multirow{2}{*}{MT06}}  & \multicolumn{4}{c||}{MT08} & \multirow{2}{*}{avg.}\\
                                      &              &  & & \multicolumn{2}{c|}{News} & \multicolumn{2}{c||}{Web} &  \\\cline{3-9}
                                      &              & \bp & \tb & \bp & \tb & \bp & \tb & \tb \\ 
\hline
\multicolumn{2}{c||}{Tree-to-string} & 0.90 & 8.70 & 0.84 & 12.65 & 0.84 & 17.00 & 12.78 \\
\hline \hline
\multicolumn{2}{c||}{LVNMT}          & 0.96 & 9.78 & 0.94 & 14.15 & 0.97 & 15.89 & 13.27 \\
\hline
  &  {\bf U$_{GRU}$}  & 0.97 &  8.62 & 0.95 &  12.79 & 0.97 & 15.34 & 12.31\\
\hline
\end{tabular}
\caption{Single system results in terms of (\ter-\bleu)/2 
on 11 million set. 
NMT results are on a large vocabulary ($500k$) and with UNK replaced. \label{tab:11m}
Due to the time limitation, we only have the results of {\bf U$_{GRU}$} system.}
\vspace{-0.3cm}
\end{table}

Table~\ref{tab:11m} shows the results of 11 million systems, 
LVNMT achieves an average (\ter-\bleu)/2 of 13.27, which is about 2.5 points better than 5 million LVNMT.
The result of our {\bf U$_{GRU}$} coverage model 
gives almost 1 point gain over LVNMT.
Those results suggest that
the more training data we use, the stronger the baseline system becomes, and the harder to get improvements.
In order to get a reasonable or strong NMT system, we have to conduct experiments over a large-scale training set.

\subsection{Alignment Results}
Table~\ref{tab:align} shows the F1 scores on the alignment test set (447 hand aligned sentences).
The MaxEnt model is trained on $67k$ hand-aligned data, and achieves an F1 score of 75.96.
For NMT systems, we dump alignment matrixes, 
then, for each target word
we only add the highest probability link if it is higher than 0.2.
Results show that our best coverage model, {\bf U$_{GRU}$ + U$_{Sub}$},
improves the F1 score by 2.2 points over the sorce of LVNMT.

We also check the repetition statistics of NMT outputs.
We simply compute the number of repeated phrases (length longer or equal than 4 words) for each sentence.
On MT06 test set, the 5 million LVNMT has 209 repeated phrases,
our {\bf U$_{GRU}$} system reduces it significantly to 79,
{\bf U$_{GRU} + $U$_{Sub}$} and {\bf +Obj.} only have 50 and 47 repeated phrases, respectively.
The 11 million LVNMT gets 115 repeated phrases, and 
{\bf U$_{GRU}$} reduces it further down to 16.
Those trends hold across other test sets.
Those statistics show that a larger training set or
coverage embedding models alleviate the repeating problem in NMT.

\begin{table}
\centering
\begin{tabular}{cc||c|c|c}
\multicolumn{2}{c||}{system}  & pre. & rec. & F1 \\
\hline
\multicolumn{2}{c||}{MaxEnt} & 74.86 & 77.10 & 75.96 \\
\hline \hline
\multicolumn{2}{c||}{LVNMT}  & 47.88 & 41.06 & 44.21 \\
\hline
\parbox[t]{2mm}{\multirow{4}{*}{\rotatebox[origin=c]{90}{Ours}}}  & {\bf U$_{GRU}$}   & 51.11 & 41.42 & 45.76 \\
 & {\bf U$_{Sub}$}                 & 49.07 & 42.49 & 45.55 \\
 & {\bf U$_{GRU}$}+{\bf U$_{Sub}$} & 49.46 & 43.83 & 46.47 \\
 & +{\bf Obj.}                     & 49.78 & 41.73 & 45.40 \\
\hline
\end{tabular}
\caption{Alignment F1 scores of different models. \label{tab:align}}
\vspace{-0.5cm}
\end{table}

\section{Conclusion}
In this paper, we propose simple, yet effective, coverage embedding models 
for attention-based NMT. Our model learns a special coverage embedding vector for 
each source word to start with, and keeps updating those coverage embeddings 
with neural networks as the translation goes. 
Experiments on the large-scale Chinese-to-English task
show significant improvements over the strong LVNMT system.

\section*{Acknowledgment}
We thank the anonymous reviewers for their useful comments.

\balance
\bibliographystyle{naaclhlt2015}
\bibliography{naaclhlt2015}

\end{document}